\documentclass{article}
\usepackage{latexsym}
\usepackage{graphicx}
\usepackage{xspace}
\usepackage{amsmath,amssymb}

\usepackage{algpseudocode}
\usepackage{algorithm}

\usepackage{listings}
\makeatletter
\lst@AddToHook{OnEmptyLine}{\addtocounter{lstnumber}{-1}}
\makeatother
\lstdefinelanguage{zinc} 
  {morekeywords={include,annotation,predicate,set,of,array,function,let,in,constraint,var,int,float,ann,solve,satisfy,maximize,minimize,intersect,subset,union,exists,forall,return,set_in,set_subset,string,output,show}
  ,classoffset=1
  ,sensitive=false
  ,comment=[l]{\%}
  ,morecomment=[l]{//}
  ,morecomment=[s]{/*}{*/}
  ,morestring=[b]"
  ,literate=
  }

\usepackage{tikz}
\usetikzlibrary{shapes,arrows}
\usetikzlibrary{positioning}

\newtheorem{exampleenv}{Example}

\newcommand{\xsolve}{\textsf{Xsolve}\xspace}
\newcommand{\xlearn}{\textsf{XLearn}\xspace}

\newcommand{\sol}{\textsf{Solutions}\xspace}
\newcommand{\pat}{\textsf{Patterns}\xspace}
\newcommand{\obs}{\textsf{Observations}\xspace}

\newcommand{\mltcp}{\textsf{ML-to-CP}\xspace}
\newcommand{\cptml}{\textsf{CP-to-ML}\xspace}
\newcommand{\wtml}{\textsf{World-to-ML}\xspace}
\newcommand{\cptw}{\textsf{Apply-to-World}\xspace}
\newcommand{\wtcp}{\textsf{World-to-CP}\xspace}

\newcommand{\evalw}{\ensuremath{eval\_world}\xspace}

\usepackage{authblk}

\title{The Inductive Constraint Programming Loop}
\author[1]{Christian Bessiere}
\author[2]{Luc De Raedt}
\author[2]{Tias Guns}
\author[3]{Lars Kotthoff}
\author[4]{Mirco Nanni}
\author[2]{Siegfried Nijssen}
\author[3]{Barry O'Sullivan}
\author[1]{Anastasia Paparrizou}
\author[4]{Dino Pedreschi}
\author[3]{Helmut Simonis}
\affil[1]{CNRS, University of Montpellier, France}
\affil[2]{DTAI, KU Leuven, Belgium}
\affil[3]{Insight, University College Cork, Ireland}
\affil[4]{University of Pisa, Italy}
\date{}

\begin{document}
\maketitle

\begin{abstract}
Constraint programming is used for a variety of real-world optimisation problems, such as planning, scheduling and resource allocation 
problems.  At the same time, one continuously 
gathers vast amounts of data about these problems. Current 
constraint programming software does not exploit such data 
to update schedules, resources and plans. 
We propose a new framework, that we call the 
\emph{Inductive Constraint Programming loop}. In this approach 
data is gathered  and analyzed systematically, in order 
to dynamically revise and adapt constraints and optimization 
criteria. 
Inductive Constraint Programming aims at bridging the gap 
between the areas of data mining and machine 
learning on the one hand, and constraint programming on the other hand.
\end{abstract}

\section{Introduction}

Machine Learning/Data Mining (ML/DM) 
and Constraint Programming (CP)  are  central to 
many application problems. 
ML is concerned 
with learning functions/patterns characterizing some training data 
whereas CP is concerned with finding solutions 
to problems subject to constraints and possibly an optimization function.

The problem with current  technology is that the problems of
data analysis and constraint satisfaction/optimization have
almost always been studied independently and  in isolation.
Indeed, there exist a wide variety of successful approaches to analysing
data in the field of ML{}, DM{} and statistics,
and at the same time, advanced techniques for addressing constraint
satisfaction and optimization problems have been developed in the
  {CP} community.
Over the past decade a limited number of isolated studies on specific
cases has indicated that significant benefits can be obtained by
connecting these two fields \cite{ 
EpsteinF01, XuHHL08,  RaedtGN08,BessiereHO09,KhiariBC10,CoqueryJSS12},
but so far a truly general, integrated and cross-disciplinary approach
is missing. 

CP technology is used to solve many types of problems, such as power 
companies generating and distributing electricity, hospitals
planning their surgeries, and public transportation companies scheduling buses.
Despite the availability of effective and scalable  solvers, 
current approaches are still unsatisfactory. 
The reason is when using CP technology to solve these applications, 
the constraints and criteria,  that is, the {\em model}, must be statically 
specified. 
However, in reality often this model needs to be revised over time. 
The revision can be needed to reflect changes in the environment due 
to external events that impact the problem. 
The revision can also be needed  because the execution of the solution 
generated by the  model has modified the characteristics of the problem. 
Finally the revision can be needed simply because the original 
model did not capture correctly the problem. Observing the impact of the 
solution allows us to correct or improve the model. 
Therefore,
there is an urgent need for improving and revising a model over
time {\it based on  data} that is continuously gathered about the
performance of the solutions and the environment they are used in. 

Exploiting gathered data to modify the model
is difficult and labor intensive with state-of-the-art solvers, as these 
solvers do not support DM and ML.
As a consequence, the data that is  being gathered  today in order to
monitor the quality of the produced solutions and to help evaluating the
effect of possible adjustments to the constraints or optimization
criteria, is not fully exploited when changes in a schedule or plan are
needed.
Hence,  schedules and plans that are produced are 
often  suboptimal. This, in turn, leads to a waste of resources.
Instead of using data passively, data should be actively analysed in
order to discover and update the underlying regularities, constraints
and criteria that govern the data.

In this paper, we  propose and formalize the new framework of 
{inductive constraint programming}. 
This framework is based
on what we call the Inductive Constraint Programming loop, which is an 
interaction between a machine learning component (ML) and a constraint 
programming component (CP). 
The ML observes the world and extracts patterns. The CP solves a
constraint satisfaction or optimization problem using these patterns and whose solution is applied to the world. We assume the world changes over time, possibly due to the impact of applying our solution. This process is repeated in a loop.
Inductive constraint programming will serve the
long-term vision of easier-to-use and more effective tools for resource
optimization and task scheduling.

\section{Background}\label{sec:background}

In this section we introduce the basic concepts 
used later in the paper. We briefly define and 
explain what is a constraint problem and a learning 
problem. 

\subsection{Constraint problem}\label{sec:cp}

The central notion in constraint programming is the 
\emph{constraint}. A constraint is a Boolean function whose scope 
is a set of (integer) variables. Depending on whether the function 
returns true or false for a given input assignment of its variables, 
the constraint accepts or rejects the assignment. 
For instance, the constraint $X_1+X_2=X_3$ specifies that any 
combination of values for variables $X_1, X_2$ and $X_3$ has to 
be such that the sum of $X_1$ and $X_2$ equals $X_3$. 
Based on the notion of constraint, we define constraint network 
and solver.

A \emph{constraint network} $N=(X,D,C,f)$ is composed of: 
a set  $X$  of variables taking values in
domain $D$. These
variables are subject to constraints in the set $C$. 
The optional evaluation function $f$ takes as
input an assignment on $X$ and returns a cost for it.
A  solution (optionally \emph{best} solution) of
$N$ is a tuple in $D^X$ satisfying all the constraints in $C$ (optionally minimizing $f$). 
A \emph{solver} takes as input a constraint network and returns a 
solution/best solution or  failure  in case no solution satisfying 
all the constraints exists. 

There exist several languages/formats for specifying a constraint 
problem to be given to a solver for solving.
Take for instance the Sudoku problem. 
Figure \ref{fig:cp-example} expresses Sudoku as a constraint satisfaction 
problem, using a pseudo-MiniZinc language \cite{minizinc}. 
Line \verb|1| defines  an input 
matrix \texttt{start} containing the prefilled cells of the Sudoku. 
Line \verb|2| defines the matrix \texttt{puzzle} of 
variables that will contain the solution of the Sudoku. 
Lines \verb|4-5| put equality constraints between the prefilled cells in 
the input  matrix \texttt{start} and the matrix of variables \texttt{puzzle}. 
Lines \verb|7-8| post an  \texttt{alldifferent} constraint on every row of \texttt{puzzle}. 
\texttt{alldifferent(xi | i in 1..n)} is a global constraint that specifies 
that variables \texttt{x1..xn} must all take different values. 
Lines \verb|10-11| do the same for the columns. 
Lines \verb|13-15| is a bit more tricky as it has to play with the indices 
of the subsquares to post the \texttt{allfdifferent} constraints on the 
variables of every subsquare in \texttt{puzzle}. 
Finally, line \verb|16| calls the solver on the instance. 

\begin{figure}[tbp]
\lstset{language=zinc,escapeinside={@}{@},frame=trbl,numberblanklines=false,rulesepcolor=\color{black},numbers=left}
\begin{lstlisting}
array[1..9,1..9] of 0..9: start; %% initial board 0 = empty
array[1..9,1..9] of var 1..9: puzzle;
   
% fill initial board
constraint forall(i,j in 1..9 where start[i,j] > 0)(
  puzzle[i,j] = start[i,j] );
   
% All different in rows 
constraint forall (i in 1..9) (
  alldifferent( [ puzzle[i,j] | j in 1..9 ]) ); 
   
% All different in columns.
constraint forall (j in 1..9) (
  alldifferent( [ puzzle[i,j] | i in 1..9 ]) ); 
   
% All different in sub-squares:
constraint forall (i, j in 1..3)(
  let { int: a = (i-1)*3; int: b = (j-1)*3} in
  alldifferent( [ puzzle[a+i1, b+j1] | i1,j1 in 1..3 ] ) );
   
solve satisfy;
\end{lstlisting}
   \caption{Sudoku in  pseudo-Minizinc.   \label{fig:cp-example}}
   \end{figure}

\subsection{Learning problem}\label{sec:cp}
In machine learning, the goal is to learn a hypothesis that explains the observed data. The data typically consists of a set of training examples $E$, which are assumed to be independent and identically distributed. Different learning methods differ largely in the type of examples to learn from, and the type of hypothesis they want to learn. The most popular learning setting is supervised learning, where each example in $E$ is accompanied by a label that should be predicted. One can then search for a linear function over the examples that best predicts the labels, or for a \textit{decision tree} that does so. More formally, we define the learning task as follows:

A \emph{learning problem} $L = (E, H, t, loss)$
is composed of a set $E$ of examples, a 
hypothesis space $H$, the 
target function $t$ that one wants to
learn, and a loss function $loss(E,h,t)$ that measures
the quality of a hypothesis $h \in H$
w.r.t. dataset $E$ and the target hypothesis $t$. The task is to find a hypothesis that minimizes the loss.

For example, given real-valued data $E \subset \mathbb{R}^d$ and 
real-valued labels identified by target function $t$, where 
$\forall {\bf e} \in E: t({\bf e}) \in \mathbb{R}$, the goal of linear 
regression is to learn a linear function $h_{\bf c}: E \rightarrow \mathbb{R}$ 
with coefficients ${\bf c}$ that minimizes the sum of squared errors between 
the predicted value and the observed value: 
$loss(E,h_{\bf c},t) = \sum_{{\bf e} \in E} |h_{\bf c}({\bf e}) - t({\bf e})|^2 
= \sum_{{\bf e} \in E} |{\bf e}\cdot {\bf c} - t({\bf e})|^2$. 
Many other loss functions and hypothesis spaces have been defined in the literature.

A range of machine learning methods such as (linear) regression and support vector machines can be expressed as standard optimisation problems (often unconstrained), where the goal is to find an assignment to function parameters such that the loss is minimized. In practice, usually specialised solving methods are used.


\section{ 
Inductive Constraint Programming  Loop} \label{sec:iconloop}

\begin{figure}
    \begin{center}
     	\includegraphics[width=5in]{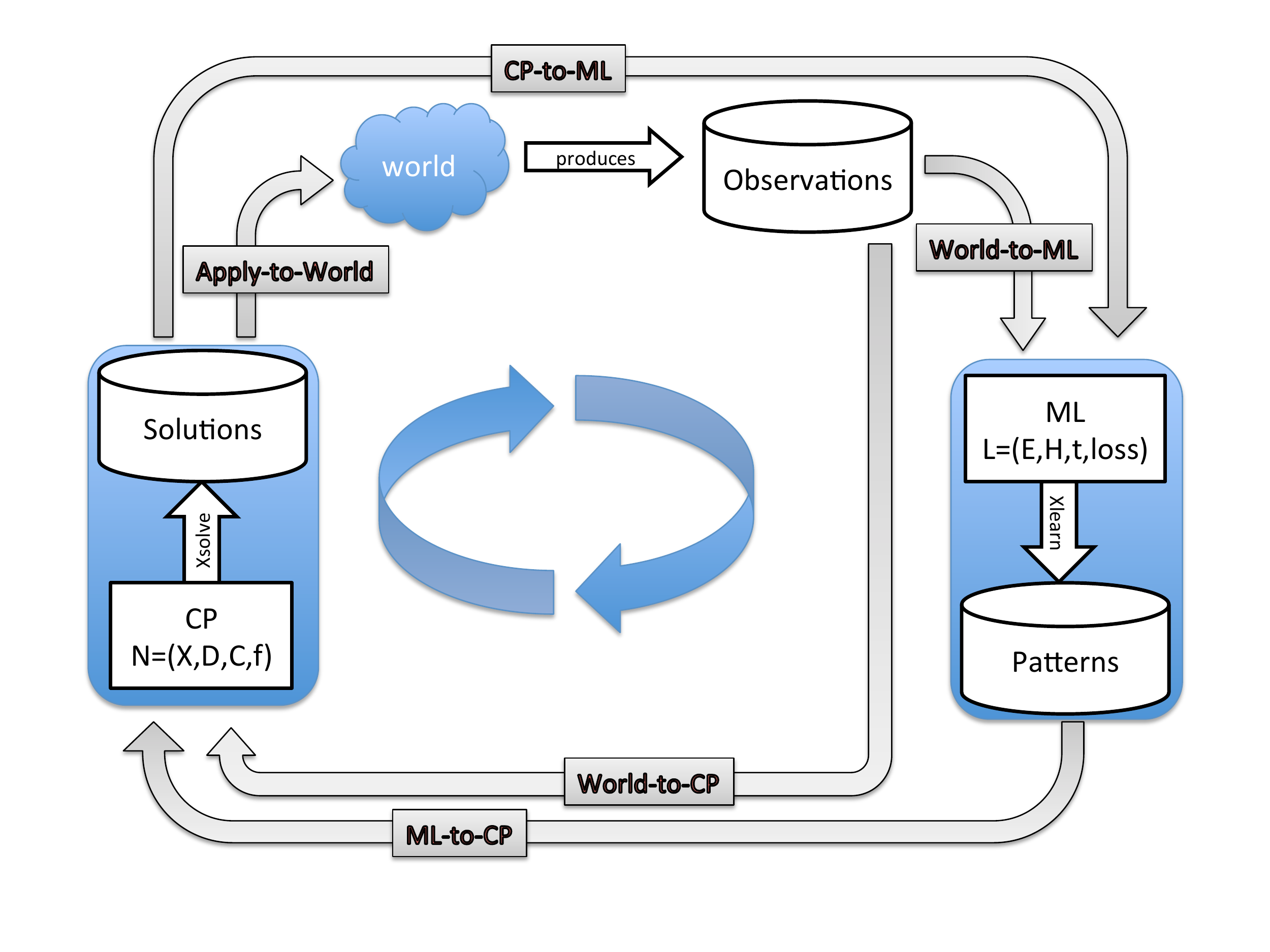}
    \end{center}
 \caption{The Inductive Constraint Programming loop \label{fig:loop}}
    \end{figure}

The inductive constraint programming loop will cope with changes in the world by iteratively solving a learning problem and a constraint problem. The loop is composed of
several components that interact with each other through writing and reading operations. 
A visualization of the loop  is given in Figure~\ref{fig:loop}.
We introduce each of the elements in the loop in turn.

The CP component is composed of a constraint network 
$N=(X,D,C,f)$ ($f$ is optional), 
a constraint solver \xsolve, and a \sol repository. 
\xsolve generates solutions  of $N$, or good/best solutions of
$N$ according to $f$,  that it writes in the \sol repository.  
In case \xsolve is not able to produce any  solution to be applied 
to the world, the CP component notifies the ML component by sending  
information about the failure. 

The ML component is composed of a
learning problem $L = (E, H, t, loss)$, a learner \xlearn,  
and a \pat repository. 
\xlearn learns hypotheses $t$ (typically one)
and writes them in the \pat repository. 

The World  component is composed of a world $W$,  
an evaluation function \evalw, and a \obs repository. 
The world $W$  can have its own independent behavior, 
dynamically changing  under the effect of time and the effect of  
applying solutions of the \sol repository. 
The solutions are evaluated by the \evalw function 
and this feedback is stored in the \obs repository. 

Now that we have defined the basis of the 
inductive constraint programming loop, we need to
define the way the CP component, the ML component and the world
interact with each other. They interact through a set of reading/writing 
functions.

An \emph{inductive constraint programming loop}  is composed 
of a world $(W,\evalw)$, a CP component
$(N,\xsolve)$, and  an ML component $(L,\xlearn)$. 
The loop uses the following channels of communication:
\begin{itemize}
\item  function \wtml reads data and evaluations from the \obs repository and
updates the learning problem $L$, that will be used by  \xlearn to learn a hypothesis $h$;   
\item  function \cptml is used to send feedback from the CP component 
to the ML component when \xsolve cannot find  any satisfactory solution 
to be applied to the world;
\item function \wtcp reads data from the \obs repository that can be 
used to directly update the constraint network $N$ used by \xsolve;  
\item function \mltcp  reads patterns from the  \pat repository and updates 
the constraint network $N$ used by \xsolve to produce solutions; 
\item function \cptw takes solutions in the \sol repository and applies them 
to the world, if possible. 
\end{itemize}

The following pseudo code demonstrates how these communication channels are used in the inductive constraint programming loop:

\begin{algorithm}[H]
\caption{Pseudo code of a loop cycle using the components.}
\begin{algorithmic}
\Function{cycle}{Observations, \textit{optional} PreviousSolutions}
 \State $L_o \gets$ \textsf{World-to-ML}(Observations)
 \State $L_p \gets$ \textsf{CP-to-ML}(PreviousSolutions)
 \State $L \gets$ constructL($L_o,L_p$)
 \State Patterns $\gets$ applyXlearn($L$)
\vspace{4px}
 \State $N_o \gets$ \textsf{World-to-CP}(Observations)
 \State $N_p \gets$ \textsf{ML-to-CP}(Patterns)
 \State $N \gets$ constructN($N_o,N_p$)
 \State Solutions $\gets$ applyXsolve($N$)
\vspace{4px}
 \If{\textsf{Apply-to-World}(Solutions)}
 \State \Return 
 \Else
 \State cycle(Observations, Solutions)
 \EndIf
 \EndFunction
\end{algorithmic}
\end{algorithm}

Initially, \wtml is used to gather training data to the ML component. 
These data can be feedback from previous executions of solutions of the 
CP component on the world. 
The solution of the previous cycle can also directly be used as well, through \cptml. This is especially useful if the previous solution could not be applied to the world, for example because the learned patterns lead to an inconsistency. Using the output of \wtml and \cptml, the learning problem $L$ can then be constructed, specific to the learner at hand. Next, the learner is applied to $L$ and patterns are obtained. These patterns can be weights of an objective function, constraints, or any other type of structural information that is part of the CP problem.

A similar process then happens for the CP component, the network is constructed using the output of \wtcp and \mltcp, after which the solving method is used and solutions are obtained.

These solutions are then applied to the world using \cptw. As mentioned before, it may be that the found solution (or non-solution) is not applicable to the world. In that case, a new iteration of the loop is started immediately which bypasses the world. Otherwise the solutions are applied to the world, after which a new cycle with new observations can be started.

We can observe that there is no direct link between the ML component 
and the world. 
Our framework is indeed devoted to solving combinatorial problems such as scheduling and routing, revising them based on 
feedback from the world; it does not aim to only classify or predict events in the world.

\section{Illustrative Example}\label{sec:showcase}
To illustrate the inductive constraint programming loop we will 
use a scheduling setting that occurs in hospitals. 
This setting includes an ML component, a CP component and a world component.

We will first describe the CP component. In this component we focus 
on a task scheduling problem. The treatment of a patient typically 
involves the execution of various tasks on this patient, such as 
executing scans, taking blood tests, operating the patient, physiotherapeutic sessions, and so on. These tasks need to be executed in a well-defined order, and require the use of the resources of the hospital for a certain amount of time. The overall scheduling problem is how to schedule these tasks in the shortest amount of time possible, using the limited resources of the hospital.

Important parameters of this scheduling problem hence include the 
resources available in the hospital and the tasks that need to be 
executed. For each task, it is important which resources need to be 
used, how many such resources are needed, and for how long they need 
to be used.

Whereas for many patients it is clear which procedures need to be followed before the patient can be discharged from the hospital, this is not the case for the duration of these tasks: depending on parameters such as age or health conditions, a certain task may take much longer for one patient than for another patient. 

The task of the ML component is to address this challenge: its task is to predict how long a task is estimated to take for a patient. This task involves solving a regression problem as identified earlier: for each given task for a patient, we need to predict its duration, which is a real number.

The world component executes the schedules; it produces data about patients and observations concerning the true durations of tasks.

Clearly, as the tasks are executed in the hospital, the predicted durations may differ from the actual durations. Furthermore, new  patients and hence new 
tasks arrive. This means that the hospital needs to schedule 
tasks on a regular basis. 
The patient data that is collected during each such iteration 
can here be used to improve the quality of the predicted task durations. 
This makes it a good example of the inductive constraint programming loop.
Within this loop, we can distinguish the following components and functions:
\begin{itemize}
\item function \wtml reads historical patient data and historical task durations for these patients; furthermore, it reads the patients that are currently in the hospital and the tasks that need to be executed for these patients;
\item the ML component  predicts the durations for the tasks that need to be executed, using the historical data;
\item function \mltcp reads the learned durations and updates the CP network accordingly; 
\item function \wtcp reads the tasks that need to be executed from the world, as well as the resources available in the hospital; 
\item the CP component solves the updated scheduling problem; 
\item function \cptw applies the resulting schedule in the world.
\end{itemize}
In this example, the function \cptml is not used; it could be used, for instance, if there is a preference to schedule nurses and doctors in similar teams or with similar load or time-breaks from day-to-day. 

Both components can be formalized using a CP language, such as the Mini\-Zinc 
language mentioned earlier. Figure~\ref{fig:cp-hospital-scheduling} shows 
MiniZinc code for the task scheduling problem. 
In this model, the parameters of the problem setting are reflected as follows:
\begin{itemize}
\item the \verb|dur| array represents the durations of all the tasks, 
as predicted by the ML component (line \verb|3|);
\item the \verb|prev| array indicates for each task which task needs 
to be executed before this task; note that we assume that there is a 
dummy first task that precedes all tasks (line \verb|4|);
\item the \verb|cap| array represents the capacity of the resources available 
(line \verb|7|);
\item the \verb|use| array represents how many resources of each type need to 
be used to execute a certain task (line \verb|8|).
\end{itemize}

The variables that need to be found are the \verb|start| variables 
(line \verb|11|), which indicate at which times the tasks need to be 
executed. The constant  \verb|max_time| represents the latest time at which a task may still start, this could be specified for each task separately as well.

The constraints are twofold:
\begin{itemize}
\item the constraint on line \verb|14| is a \verb|cumulative| 
constraint; for a given resource, it ensures for each time point 
that the use of the resource is within the capacity bound of that 
resource. Note that the \verb|cumulative| constraint is a built-in 
constraint available in the MiniZinc language. Constraints that can involve any number of variables are called \emph{global} constraints. They can capture complex structural constraints of the problem. Global constraints are an essential part of the efficiency of CP models. 
\item the constraint on line \verb|17| ensures that a task only executes after the task that should precede it has finished.
\end{itemize}
The optimization criterion is to minimize the makespan, that is, 
to assign the \verb|start| variables so that the total amount of time 
used by the schedule is minimum (line \verb|19|). 

\begin{figure}[tbp]
\lstset{language=zinc,escapeinside={@}{@},frame=trbl,numberblanklines=false,rulesepcolor=\color{black},numbers=left}
\begin{lstlisting}
% Tasks: duration and precedence
int: nbTasks; set of int: Tasks = 1..nbTasks;
array[Tasks] of int: dur;
array[Tasks] of int: prev;

% Resources: capacity and use
int: nbRes; set of int: Res = 1..nbRes;
array[Res] of int: cap;
array[Res,Task] of int: use;

% Variables: start times
int: max_time;
array[Tasks] of var 0..max_time: start;

% Resource capacities
constraint forall(r in Res) (
  cumulative(start, dur, use[r], cap[r]) );

% Precedence between tasks
constraint forall(t in Tasks) (
  start[t] > (start[prev[t]] + dur[prev[t]]) ));

% Minimize the amount of time
solve minimize max(t in Tasks) (start[t]+dur[t]);
\end{lstlisting}
   \caption{The hospital scheduling problem in pseudo-Minizinc.  \label{fig:cp-hospital-scheduling}}
   \end{figure}

\begin{figure}[tbp]
\lstset{language=zinc,escapeinside={@}{@},frame=trbl,numberblanklines=false,rulesepcolor=\color{black},numbers=left}
\begin{lstlisting}
% Dimension of the input data.
int: N; % Number of observations
int: M; % Dimension of observations

% Input data: observed data (X) and target labels (Y)
array[1..N, 1..M] of float: X; 
array[1..N] of float: Y;       

% Weights to fit (W[M+1] is constant term)
array[1..M+1] of var float: W;

% Calculate predictions and errors
array[1..N] of var float: Est =       
            [ sum(j in 1..M) (W[j]*X[i,j]) + W[M+1] | i in 1..N ];
array[1..N] of var float: Err =       
            [ Est[i] - Y[i] | i in 1..N ];  

% Minimize the squared error 
solve minimize norm2(Err);

% Auxiliary functions for computing the 2-norm
function var float: norm2(array[int] of var float: W) = 
        sum(j in index_set(W)) ( W[j]*W[j]);
\end{lstlisting}
   \caption{The hospital learning problem in pseudo-Minizinc.  \label{fig:cp-hospital-learning}}
   \end{figure}

To predict the durations of the tasks in the hospital, a regression task needs to be solved.
Many different models can be made for this regression task, each corresponding to learning a different type of regression model. Arguably the most simple regression model is the linear model, in which the task duration prediction is based on a linear combination of the characteristics of the patient on which the task is executed.

The problem of learning such a regression model is formalized in Figure~\ref{fig:cp-hospital-learning}. Variables \verb|X| and \verb|Y| represent the training data, where \verb|X| contains the descriptive attributes of various tasks and \verb|Y| the historical durations of these tasks;  variable \verb|W| represents the weights of the features that we are learning.

Based on these weights, we can calculate an error for the predictions; line \verb|10| calculates a weighted linear combination for each training example, using the weights \verb|W|; this prediction is used in line \verb|12| to calculate an error for each example. Line \verb|15| minimizes the error over all examples, where line \verb|17| defines that the errors for the individual examples are combined by summing the squared errors.

The scheduling model and the machine learning model together define both components of the inductive constraint programming loop. We here demonstrated how a declarative, unified language could be used to model both the learning problem and the solving problem. While a single language for both the learning and solving is an appealing prospect, it is not a requirement for the applicability of the inductive constraint programming loop.

\section{Other Examples}

We now briefly describe a number of other problems that 
can be captured in the inductive constraint programming loop.  
For each of them, we define the problem, and then define 
the interactions between the world, the ML component, 
and the CP component.
The first three problems we describe 
(optimizing bus schedules, car pooling, and energy-aware data centers)  
are real world problems 
that can be expressed in a neat and efficient way through 
the inductive constraint programming loop.  
The two last (constraint acquisition and portfolio selection) are 
existing academic problems that can be seen with a new 
eye through the inductive constraint programming loop. 


\subsection{Optimizing bus schedules}

In order to improve human mobility,  the region 
of Pisa plans to take into account information about the trajectories 
of people taking their car in order to improve the public transportation system. 
The problem is composed of
two parts. The first one consists in tracking the GPS localisation of
cell phones to understand the way people commute in the Pisa
region. The second part consists in optimising bus schedules to
meet  as much as possible the requirements of these people. The problem
is dynamic  as
the implementation of the generated bus schedule will affect the
way people commute, which should  be observed again, and so on. 
Such a problem can be represented in the inductive constraint programming loop framework.

    \begin{figure}
    \begin{center}
     	\includegraphics[width=2.5in]{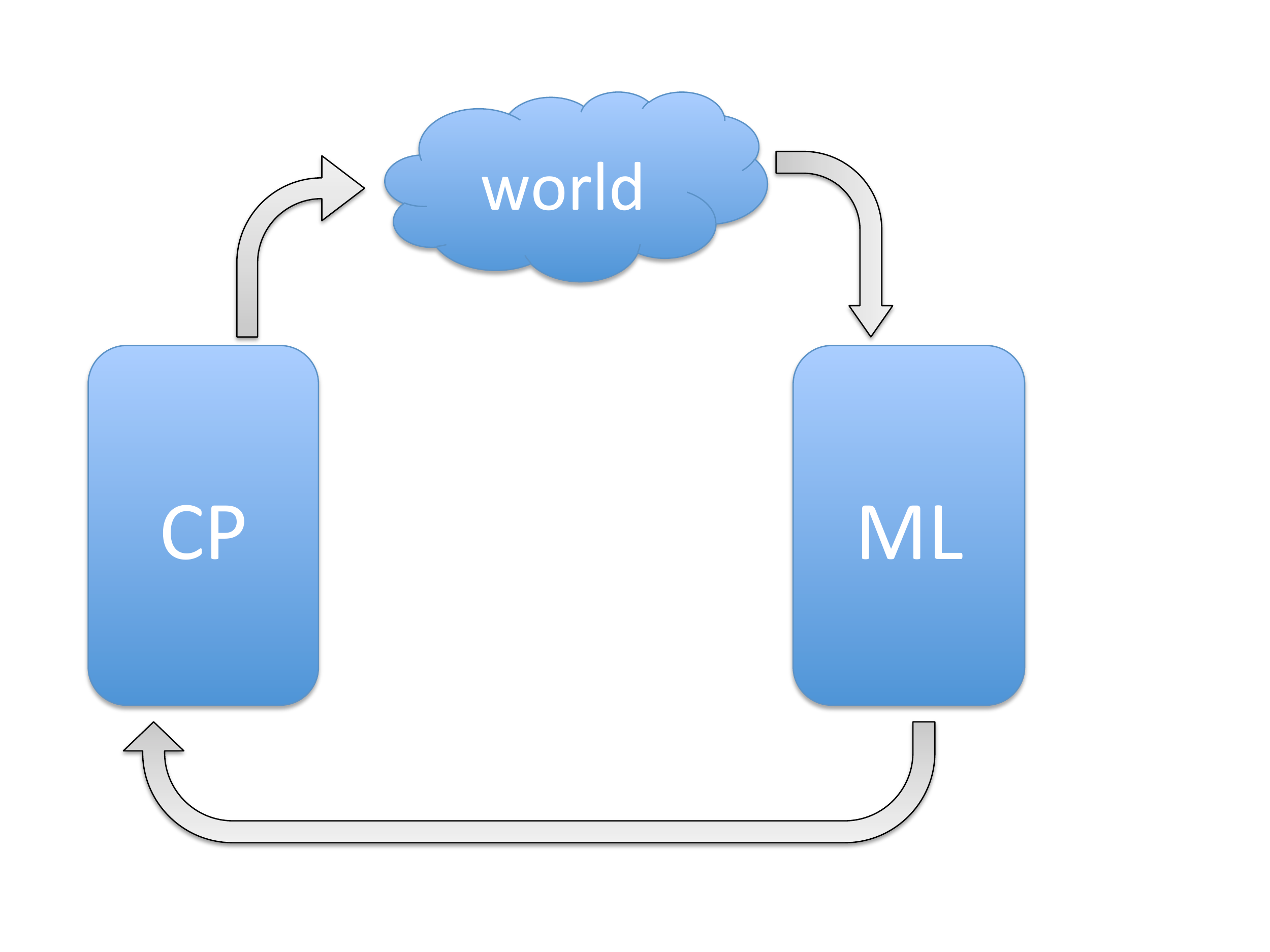}
    \end{center}
 \caption{The  loop of the bus schedule problem \label{fig:bus}}
    \end{figure}
    
The world is observed through the GPS localisation of cells phones in the
region of Pisa and is represented by a set of trajectories of individuals. 
An evaluation of the traffic quality is provided by the \evalw function. 
\evalw is based on a measure of the amount of traffic jams generated by 
the traffic. 
All these observations are stored in the \obs repository. 
The function \wtml  reads observations that are  given as input to the ML component. 
The ML component  uses these observations to learn patterns from 
the trajectories and from the quality evaluation of the  traffic. These 
learned patterns on the trajectories/time slots are written 
in the \pat repository. 
The CP component contains a constraint network that models the problem
of generating good bus schedules for the region of Pisa, that is, 
bus schedules that cover as much as possible trajectories of people 
at the time they need them. It is
parameterized by the weight of each trajectory/time slot. The values
of these parameters are computed by the function \mltcp based on the input 
from \pat.  The output of the CP 
component is a new bus schedule that is written in the \sol repository 
to be applied to the world by function \cptw. 
The process can loop for ever. 
Figure \ref{fig:bus} shows the loop solving this problem.

\subsection{Carpooling}

The carpooling application is aimed at proposing carpooling 
matches to a set $U$ of users participating to the service, 
based on their actual mobility (the trips they performed with 
their private cars) and any information about which kind of 
match proposals are likely to be accepted by a user.
This problem can be modelled in  the inductive constraint programming loop framework.  

    \begin{figure}
    \begin{center}
     	\includegraphics[width=2.5in]{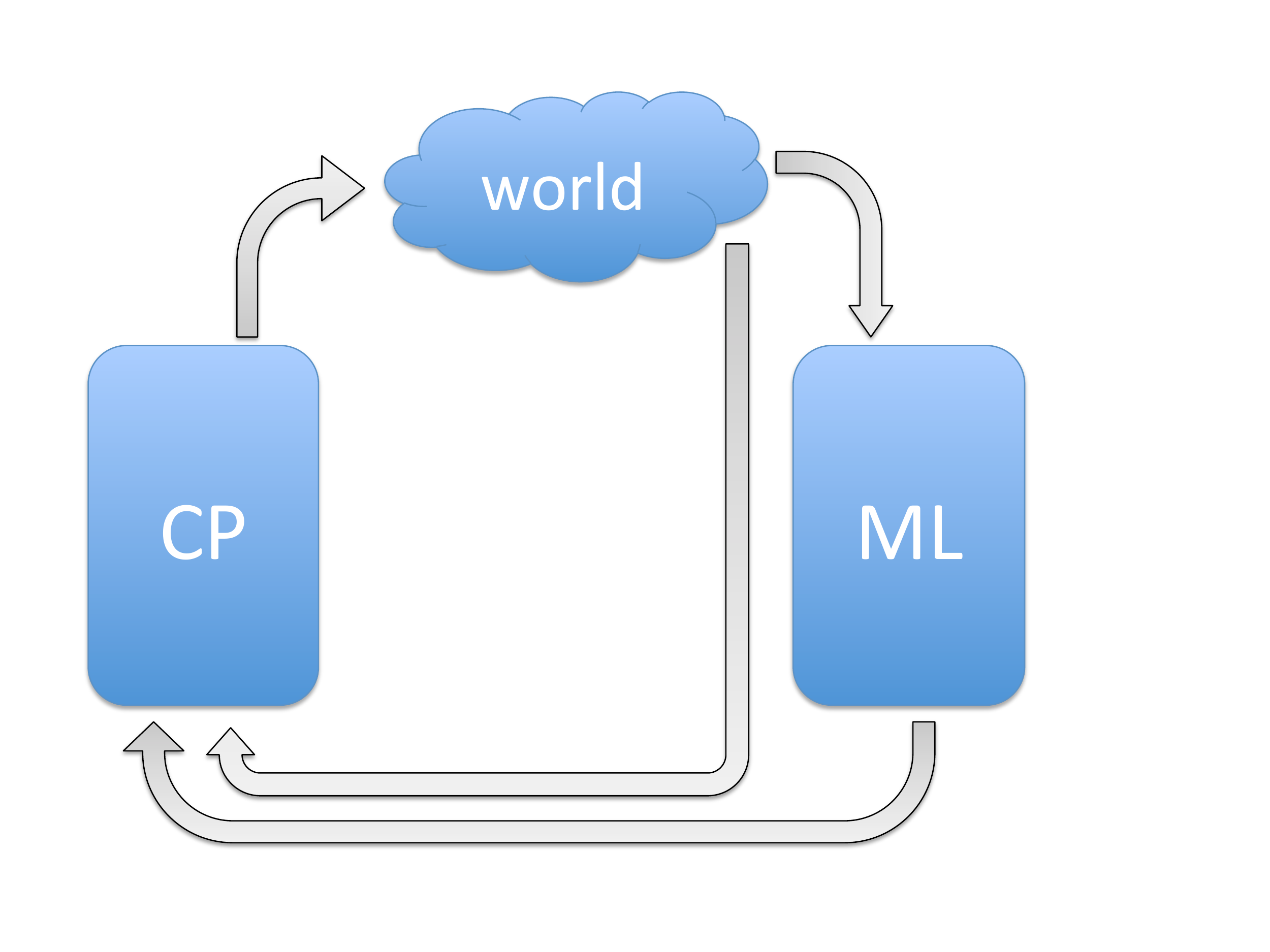}
    \end{center}
 \caption{The  loop of the carpooling problem \label{fig:carpooling}}
    \end{figure}

The world is observed through a set of raw trajectories for each user, 
describing the recent mobility of that user. 
The \evalw function returns the response of each user to the 
previous carpooling solution offered to her (accepted, rejected, 
will reject next time). All these data are written in the \obs repository. 
The function \wtml simply reads this information and sends it to the
ML component, which will perform some mining operations to produce 
a temporally labelled weighted oriented graph $G=(U,E,t,p)$, 
where there exists an edge 
$(u,v)\in  E$ if and only if user $v$ can give a lift to user $u$. 
Each edge $e$ is labelled with the time $t(e)$ the lift can take place, 
and a probability $p(e)$ of having the match proposal being accepted 
by both $u$ and $v$. 
The graph $G$ is written in the \pat repository. 
\mltcp reads this graph and encodes it as a constraint network. 
The observation of the world also provides other information 
useful to the carpooling system, e.g. 
the number of passengers that can be hosted in the user’s vehicle.
The function \wtcp reads this information 
and directly sends it to the CP component, which will add the relevant 
constraints in 
the network. For each user $u$, a constraint $c(u)$ defining 
the maximum capacity of the vehicle of user $u$ wil be added. 
The CP component solves the network to find the best solution. 
This solution is a carpooling assignment that maximizes 
the reduction of cars and the likelihood of acceptance by users. 
It is written in the \sol repository, then proposed to the users. 
Figure \ref{fig:carpooling} shows the loop solving this problem.

\subsection{Energy-aware Data Centres}

The aim is to improve the energy-efficiency of data-centers.
Consider a cloud computing service, where customers contract to run computing services (tasks) throughout the day.
Tasks are assigned to machines within the data centre and require a certain amount of resources for the duration which they run. The aim is to schedule these tasks such that the overall cost of energy used is minimized.
However this is complicated by the fact that large electricity consumers, like a data centre, will typically pay a variable price for their electricity, which is not known in advance. In Ireland for example, the price is not known until four days after. The price may also fluctuate significantly throughout the day, which provides the opportunity to reduce the energy use during peak periods and instead perform the work during cheaper periods. This requires a forecast of the price ahead of time and to produce a schedule of  the tasks based on the forecasted price data.

    \begin{figure}
    \begin{center}
     	\includegraphics[width=2.5in]{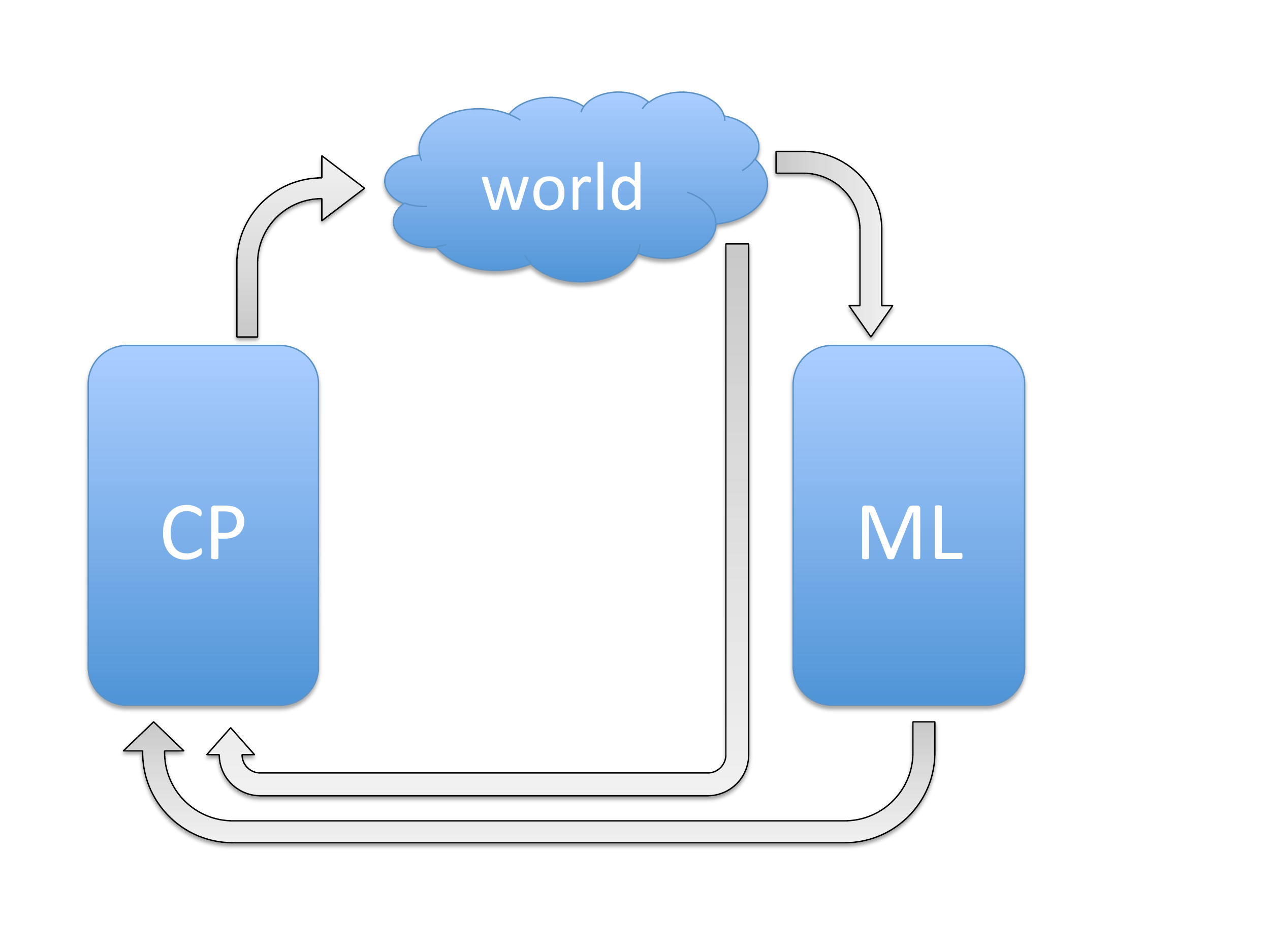}
    \end{center}
 \caption{The  loop of the energy-aware data center problem \label{fig:energy}}
    \end{figure}

This application was the focus of the first inductive CP Challenge\footnote{http://iconchallenge.insight-centre.org/} and 
is modeled in the inductive constraint programming loop as follows. 
The world consists of a number of elements: the wide range of factors 
which affect the energy market, like weather conditions, producers/consumers 
of electricity, etc; and customers of the data center who contract the 
various workloads.
\wtcp gives to the CP component the  tasks to be scheduled at the next turn;  
\wtml takes input from the world to produce a hypothesis $h$ modeling the 
electricity price.
\mltcp incorporates this forecast to produce a solution to the scheduling problem minimizing the forecast energy cost. 
\cptw takes this schedule and applies it to the world. As time 
progresses and the world changes \wtml will need to evolve the 
forecast model, and 
subsequently the schedule, to take account of factors affecting the energy price.
Figure \ref{fig:energy} shows the loop solving this problem. 

Here, the machine learning is applied once and the outcome is directly used by the CP program. An alternative, proposed by Tulabandhula and Rudin~\cite{TulabandhulaRu13}, is to do the machine learning while taking the operational cost (the outcome of the CP problem with the learned weights) into account. This can be achieved by making the operational cost a part of the loss function of the ML problem. One can then repeatedly iterate between solving the ML and CP component, before applying the found schedule in the world.


\subsection{Constraint acquisition}

Modeling a problem as a constraint network requires expertise in constraint 
programming. 
If we want novices to use constraint programming, we need automatic constraint 
acquisition systems that assist the user in the modeling task. 
CONACQ is such a system \cite{besetalIJCAI07queries}.  
CONACQ interacts with a user to learn 
a target constraint network that represents the problem of the user.
We describe how CONACQ can be implemented as an instance of the inductive constraint programming loop. 

    \begin{figure}
    \begin{center}
     	\includegraphics[width=2.5in]{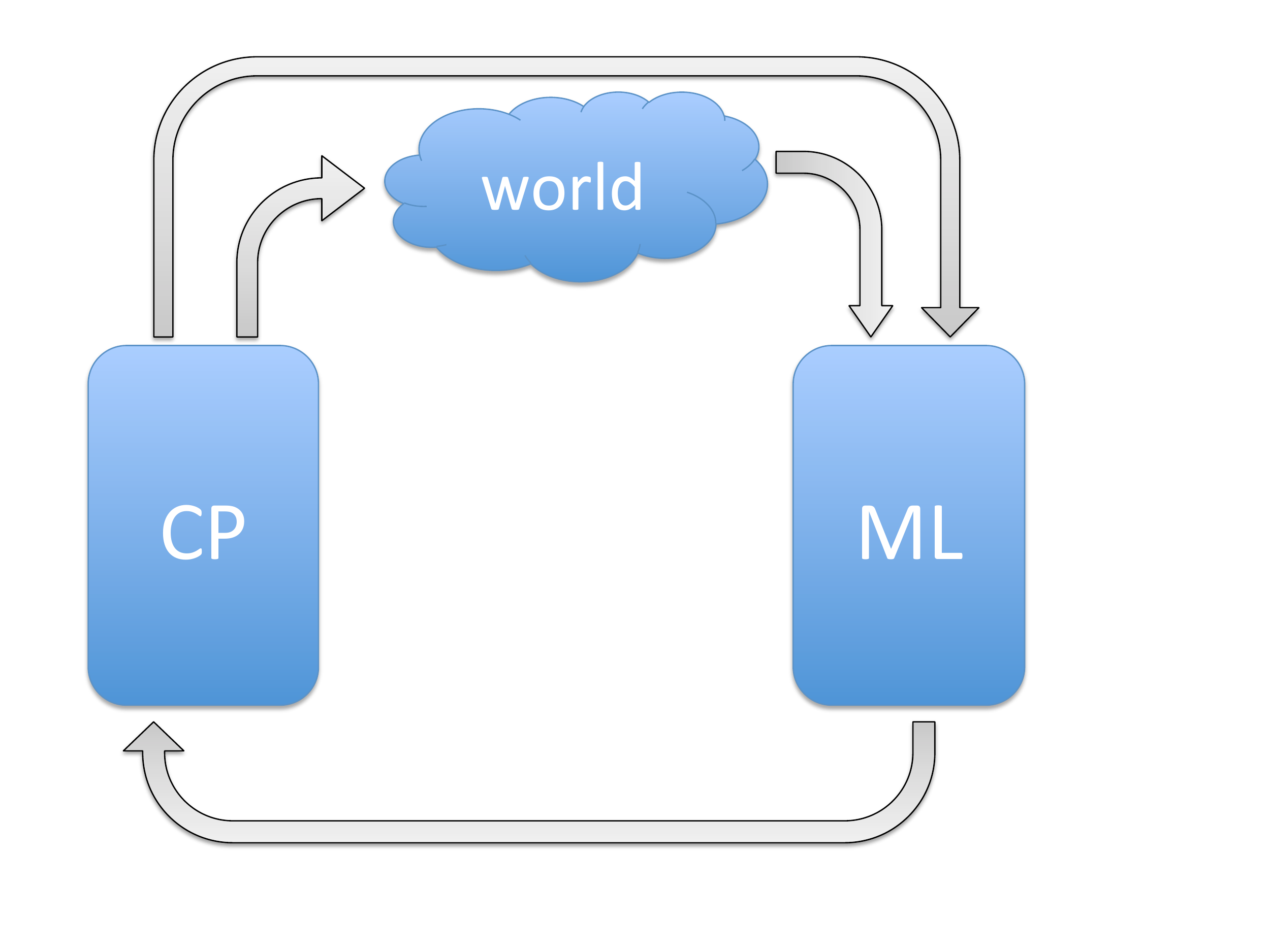}
    \end{center}
 \caption{The  loop of the constraint acquisition  problem \label{fig:conacq}}
    \end{figure}

The world involves a set of examples defined on a set of objects/variables $X$. 
An example is an assignment of a value to each of the variables in $X$. 
Examples are produced by the CP component or by the world itself. 
The evaluation function \evalw is the user herself, who evaluates the quality 
of examples. 
In CONACQ the quality is either true or false, depending on if the example 
is  a solution of the target constraint network or not. 
The examples and their evaluation are written in the \obs repository. 
\wtml simply reads the classified examples from \obs and gives them to 
the ML component. 
The hypotheses space $H$ used by the ML component is defined by the  
language of constraints used to express the target network. 
When reading examples from \obs, the ML component updates --if needed-- 
the hypothesis $h$ in the form of a set of constraints that correctly classify 
the examples. These learned constraints are written in the 
\pat repository. 
Based on the learned constraints, 
the ML component generates other constraints that implement the followed query 
strategy (e.g., near-miss). These constraints together with the learned constraints 
are sent to the CP component via the \mltcp function. 
The CP component solves the network built with all the constraints 
provided by the ML component and generates new solutions that are 
stored in the \sol repository. 
\cptw  sends the solutions from the repository to the world to be 
classified by the user/\evalw function. 
In case the constraint network has no solution that can be provided to the world, 
the \cptml function notifies the ML component that it was not able to generate 
a satisfactory query,  possibly with some reasons of failure such as an 
inconsistent set of constraints. 
Figure \ref{fig:conacq} shows the loop solving this problem.

\subsection{Portfolio selection}

An algorithm portfolio contains a number of algorithms or solvers which are all
suitable for solving the same kind of problem, but have different performance
characteristics. The aim is to, given a problem instance to solve, determine the
best solver for that particular problem, where ``best'' is defined according to
an application-specific metric. Algorithm portfolios have been shown to achieve
significant performance improvements over individual algorithms. One prominent
application area of these techniques is constraint programming. 

    \begin{figure}
    \begin{center}
     	\includegraphics[width=2.5in]{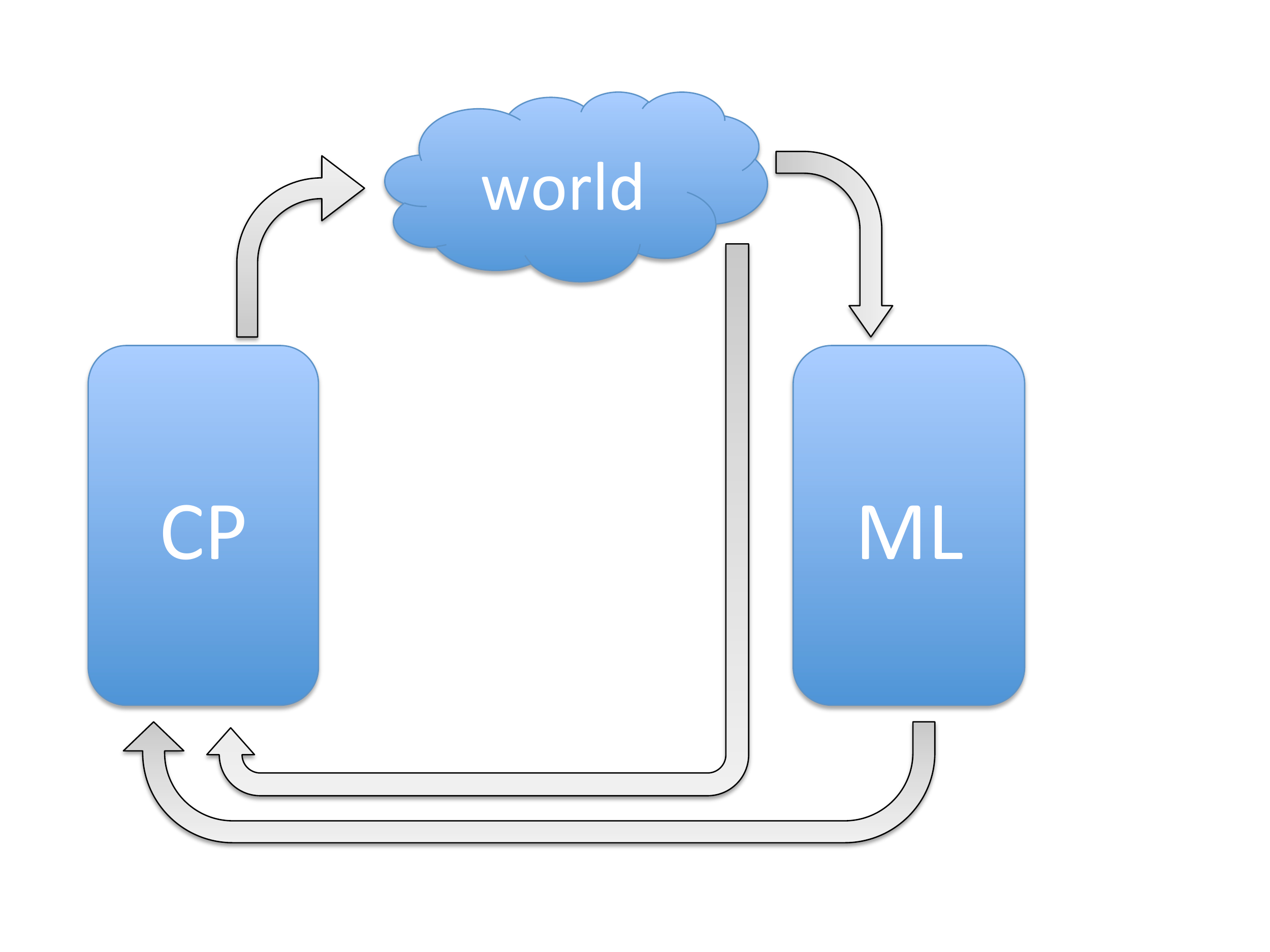}
    \end{center}
 \caption{The  loop of the portfolio selection problem \label{fig:portfolio}}
    \end{figure}

Within the inductive constraint programming loop, this problem can be modelled as a data mining 
problem that takes its data from runs of CP solvers. 
The world consists of the performance data of CP solvers on CP instances. 
\wtml reads  this performance information and the ML component learns a model 
that describes the predicted performance of CP solvers on CP instances 
and allows to determine the best CP solver for a
particular CP instance. 
When a \emph{new} CP  instance to be solved appears in the world, 
\wtml and \wtcp read it. \wtml sends it to the ML component, 
which classifies it and writes the classification in the \pat repository.  
\wtcp sends the CP instance to be solved to the CP component. 
\mltcp reads in the \pat repository 
the classification provided by the ML component to decide which solver to use. 
The  selected solver then solves
the instance and generates a new data point in the world through the \cptw function. 
This new world can lead the ML component to update the predictive model, and so forth.
Figure \ref{fig:portfolio} shows the loop solving this problem.

%

\section{Conclusion}
After a brief introduction to  constraint programming and  
machine learning, we have introduced the framework of inductive 
constraint programming. 
The key idea in the inductive constraint programming loop is that the CP and ML 
components interact with each other and with the world in 
order to adapt the solutions to changes in the world.
This is  an essential need in 
problems that change under the effect of time, or 
problems that are influenced by the application of a previous solution. It is also very effective for problems that are only partially specified and where the ML component 
learns from observation of applying a partial solution, e.g. in the case of constraint acquisition.
We have presented multiple examples of the use of 
inductive constraint programming loop 
in real world problem settings. Many other settings exist, and as more and more often learning methods are used for producing schedules and other operational plans, the need for a framework that can adapt to changes in the world will increase.

\bibliography{biblio}
\bibliographystyle{alpha}

\end{document}